\documentclass[runningheads]{llncs}
\usepackage[T1]{fontenc}
\usepackage{graphicx,verbatim}
\usepackage{tabularx}
\usepackage{multirow}
\usepackage{amsmath}
\usepackage{url}

\begin{document}
\title{TextSLIP: Text Self-Supervised CLIP for Medical Report Generation}
%

\author{Haoyu Jiang\inst{1}\thanks{Corresponding author.} \and Ziping Cong\inst{1}}
\authorrunning{H. Jiang and Z. Cong}
\institute{Centre for Artificial Intelligence and Robotics (CAIR)\\
Hong Kong Institute of Science \& Innovation\\
Chinese Academy of Sciences\\
\email{haoyu.jiang@cair-cas.org.hk}\\
\email{ziping.cong@cair-cas.org.hk}}
  
\maketitle              
\begin{abstract}
Automating radiology report generation is important for improving reporting consistency and clinical workflows . While Contrastive Language--Image Pretraining (CLIP) has advanced medical vision language modeling, existing CLIP-style approaches may still provide insufficient fine-grained semantic supervision for complex report generation. Standard CLIP primarily optimizes cross-modal alignment, without explicitly structuring the textual embedding space that guides visual representation learning. To address this limitation, we propose TextSLIP, a general medical vision-language pretraining framework that augments CLIP with intra-modal text contrastive learning. By improving textual embedding discriminability through self-supervised augmented text pairs, TextSLIP is designed to provide finer-grained linguistic supervision to the visual encoder. As an initial validation, we pretrain TextSLIP on a curated dataset of 7 million brain MRI image-text pairs and fine-tune the pretrained visual encoder within a report generation architecture. In controlled comparisons with CLIP-style baselines, TextSLIP shows consistent improvements on report generation metrics. Ablation studies further suggest that text-side self-supervision contributes to the observed gains. These results indicate that text-level contrastive learning is a promising direction for improving medical visual-textual alignment, while broader validation across additional medical domains remains an important next step. The code is publicly available at \url{https://github.com/goodrain553/TextSLIP}.


\end{abstract}

\section{Introduction}

Medical imaging reports play a pivotal role in clinical practice, as they distill complex visual findings into actionable medical knowledge that guides patient care. Automating the generation of such reports holds great promise for alleviating the workload of radiologists while improving the consistency and quality of diagnostic reporting. Radiology report generation is therefore a desirable task, aiming to automatically produce free-text descriptions from medical images.

Prior studies \cite{chen2020generatingr2gen,chen2021crossr2genCMN,wang2023metransformer} have introduced powerful transformer-based modules for radiology report generation, significantly improving the quality of generated reports. Some approaches \cite{yan2022clinical,wang2025cxpmrg,li2023dynamic} further enhance performance by optimizing the pretraining process. More recently, contrastive language–image pretraining (CLIP) has gained popularity in the development of medical foundation models \cite{khattak2024unimed,zhang2023biomedclip}, offering new opportunities to leverage large-scale image-text data for radiology report generation.

Although CLIP-based models excel at image classification, they often struggle with medical report generation. For text-oriented downstream tasks, the visual encoder must align images with fine-grained textual semantics, rather than relying solely on coarse image-text matching signals. However, standard CLIP optimizes text representations primarily through cross-modal alignment, lacking explicit supervision to structure the textual embedding space itself. Consequently, semantically similar medical descriptions with varying phrasings may not be consistently represented, weakening the fine-grained guidance provided to the visual encoder. To address this, we propose TextSLIP, a general medical vision-language pretraining framework that integrates CLIP with ESimCSE \cite{wu2022esimcse}, a sentence-level contrastive learning method. By introducing intra-modal text contrast, TextSLIP enhances the semantic discriminability of textual embeddings, enabling the visual encoder to receive finer-grained and more robust linguistic supervision.

To evaluate TextSLIP, we first pretrain the model on a dataset of 7 million brain MRI image-text pairs extracted from MedTrinity-25M \cite{xie2024medtrinity}. This brain MRI setting serves as an initial validation domain rather than a restriction of the proposed framework. To ensure fair comparison, the CLIP baseline is trained on the identical dataset, isolating the impact of the pretraining objective. For downstream assessment, the pretrained visual encoder is integrated into the R2Gen architecture \cite{chen2020generatingr2gen} and fine-tuned end-to-end. Experimental results show that TextSLIP consistently improves over CLIP-style baselines under this controlled setting. In this paper, our contributions are as follows: 
\begin{enumerate}
    \item \textbf{A text self-supervised framework for fine-grained semantic supervision.} We propose \textbf{TextSLIP}, a general medical vision-language pretraining framework that integrates contrastive language--image pretraining with intra-modal text contrastive learning. By enhancing the discriminability of textual embeddings, TextSLIP aims to provide finer-grained semantic supervision to the visual encoder for medical report generation. 
    
    \item \textbf{Large-scale initial validation on domain-specific data.} We extract approximately 7 million brain MRI image-text pairs from MedTrinity-25M \cite{xie2024medtrinity}. Utilizing this neurological corpus, we train a vision-language model under the TextSLIP framework and use it as an initial testbed for evaluating text-side self-supervision in medical report generation.

    \item \textbf{Evaluation against CLIP-style baselines.} We compare TextSLIP with strong medical visual encoders and a CLIP baseline pretrained on the same data. Ablation studies suggest that the additional text self-supervised objective contributes to consistent improvements in report generation metrics.

\end{enumerate}

\section{Method}

We introduce TextSLIP, a framework combining text self-supervision with contrastive language–image pretraining. By applying contrastive objectives to both image–text pairs and augmented textual views, this dual-contrastive strategy enhances textual discriminability, enabling the visual encoder to learn finer-grained semantic representations. The architecture and objectives are detailed in this section.

\begin{figure}[h]
\includegraphics[width=\linewidth]{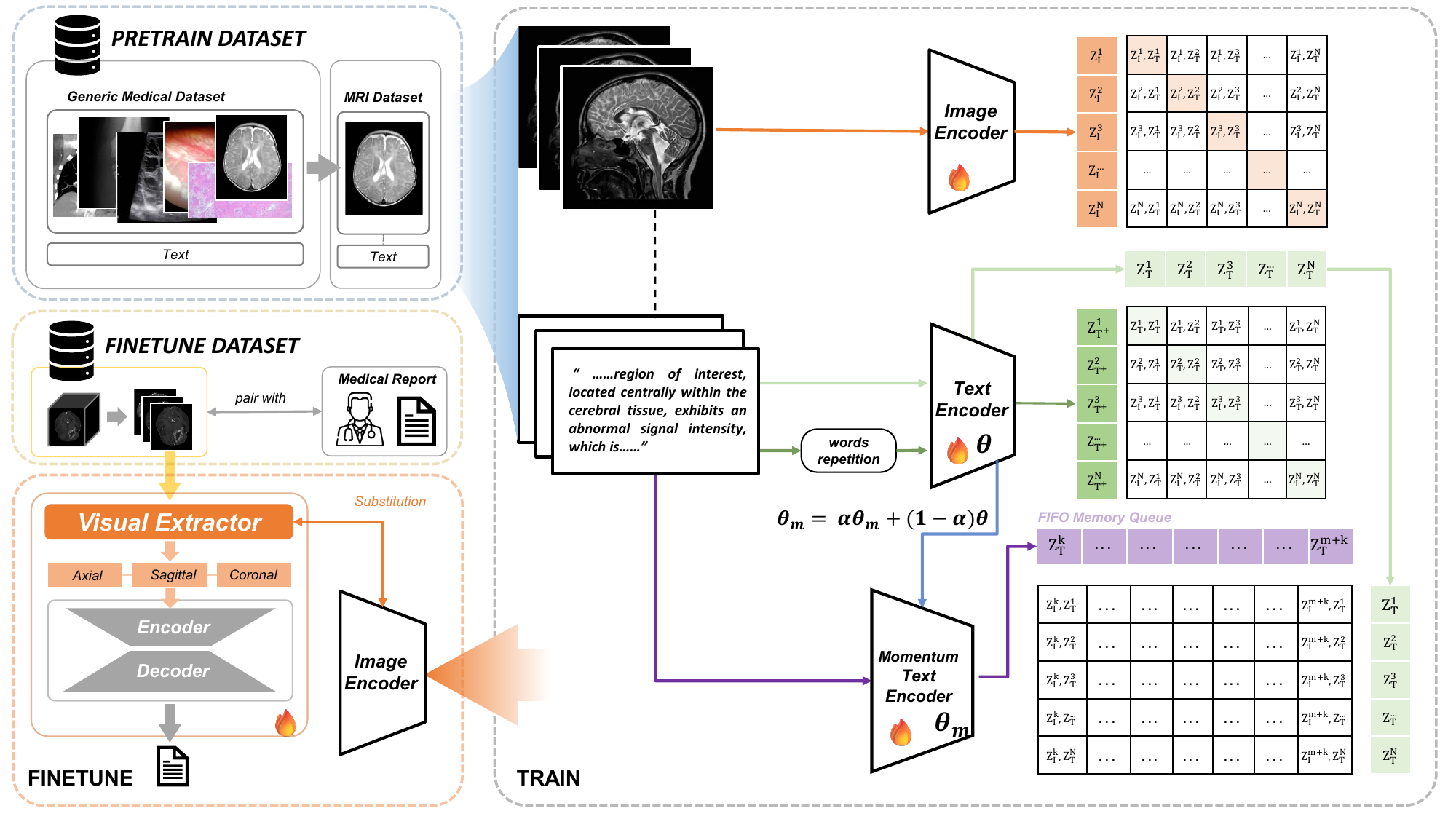}
\caption{Overview of TextSLIP} \label{fig:Overview}
\end{figure}

\subsection{Contrastive Learning Foundations}\label{subsec:CLIP}
TextSLIP builds upon two core contrastive learning paradigms: cross-modal image-text alignment and intra-modal text contrast. The following subsections detail these foundational components.

\subsubsection{Image-Text Contrastive Alignment}
Contrastive Language–Image Pretraining \cite{radford2021learningclip} learns visual representations by aligning images with corresponding captions. The framework employs modality-specific encoders to process images and text independently, projects the resulting embeddings into a shared latent space, and then normalizes them. Optimization uses the InfoNCE loss \cite{oord2018representation} to maximize cosine similarity for matched image–caption pairs while minimizing similarity for mismatched pairs within the batch.

\subsubsection{Intra-Modal Text Contrast}
Textual self-supervised learning optimizes sentence embeddings by contrasting augmented views of the same input. We adopt ESimCSE \cite{wu2022esimcse} to strengthen textual representation learning and provide robust semantic supervision for the visual encoder. Specifically, textual augmentation is achieved through a word duplication strategy, where tokens in the input sequence are randomly repeated to construct positive pairs. For negative samples, the original texts are processed through a momentum encoder, and their embeddings are stored in a first-in-first-out (FIFO) queue. The textual contrastive objective is formulated using the InfoNCE loss \cite{oord2018representation}.

\subsection{TextSLIP}

\subsubsection{Model Architecture.}

We outline TextSLIP with ESimCSE-based self-supervision in Fig.~\ref{fig:Overview}. Given a dataset $ S = \{ S_1, S_2, \dots, S_n \}$, where $ S_i = \{ T_i, I_i \}$, each image $I_i$ and text $T_i$ is mapped into a shared latent space, producing representations $z^i_I$ and $z^i_T$ through the image encoder $f_I$ and text encoder $f_T$, respectively. Additionally, a momentum encoder $ f_M$
  processes text inputs to generate embeddings $z^i_M$ 
  for constructing negative samples in the text self-supervised learning objective. The encoding process is formulated as

\begin{equation}
    z^i_I = f_I(I_i), \quad  z^i_T = f_T(T_i), \quad z^i_M = f_M(T_i)
\end{equation}

\subsubsection{Training Objectives.}

We adopt the CLIP framework \cite{radford2021learningclip} to define a bidirectional InfoNCE loss.
The image-to-text loss is computed using images as anchors.

\begin{equation}
    L_{I\to T} = -\frac{1}{N} \sum_{i=1}^N 
   \log \frac{\exp\bigl(\mathrm{sim}(z^i_I,\,z^i_T)/\tau\bigr)}
            {\sum_{j=1}^N \exp\bigl(\mathrm{sim}(z^i_I,\,z^j_T)/\tau\bigr)}
\end{equation}

Similarly, the text-to-image loss is computed using texts as anchors.

\begin{equation}
    L_{T\to I} = -\frac{1}{N} \sum_{j=1}^N 
   \log \frac{\exp\bigl(\mathrm{sim}(z^j_T,\,z^j_I)/\tau\bigr)}
            {\sum_{i=1}^N \exp\bigl(\mathrm{sim}(z^j_T,\,z^i_I)/\tau\bigr)}
\end{equation}

The CLIP objective is the average of the image-to-text and text-to-image losses, ensuring symmetric alignment between modalities.

\begin{equation}
    L^{CLIP} = \frac{1}{2}\,\bigl(L_{I\to T} + L_{T\to I}\bigr).
\end{equation}

For the text self-supervision component, we follow ESimCSE \cite{wu2022esimcse} to implement contrastive learning. Text augmentation is performed by randomly duplicating words in the input text.

\begin{equation}
T^+ = f_{WR}(T)
\end{equation}

where $f_{WR}$ denotes the word repetition operation. 

These augmented texts serve as positive samples. For negative sampling, the original texts are processed through a momentum encoder, and their embeddings are stored in a first-in-first-out (FIFO) queue. The momentum encoder parameters are updated via exponential moving average of the text encoder parameters. 

\begin{equation}
    \theta_m \leftarrow \alpha\theta_m + (1-\alpha)\theta
\end{equation}

where $\theta_m$ denotes the parameters of the momentum encoder, $\theta$ denotes the parameters of the text encoder, and $\alpha \in [0,1)$ is the momentum coefficient.

The InfoNCE loss is then computed as

\begin{equation}
L^{ESimCSE} = -\frac{1}{N} \sum_{i=1}^N 
    \log \frac{\exp\bigl(\mathrm{sim}(z^i_T,\,z^i_{T^{+}})/\tau\bigr)}
    {\sum_{j=1}^N \exp\bigl(\mathrm{sim}(z^i_T,\,z^j_{T^+})/\tau\bigr) 
    + \sum_{k \in Q} \exp\bigl(\mathrm{sim}(z^i_T,\,z^k_{M})/\tau\bigr)}
\end{equation}

where $ Q $ denotes the set of samples in the FIFO queue.
The overall training objective combines both losses with a weighting hyperparameter:

\begin{equation}
L = L^{CLIP} + \lambda L^{ESimCSE}
\end{equation}

where $ \lambda $ controls the contribution of the ESimCSE loss term.

\section{Experiments and Results}
\subsection{Datasets and Metrics}
We train TextSLIP on MRI image-text pairs and use brain MRI as the initial validation domain. Brain MRI data are extracted from MedTrinity-25M \cite{xie2024medtrinity}. BRATS2023-GLI \cite{BRATS2023-GLI} is used for fine-tuning and testing the medical report generation module. 

\subsubsection{MedTrinity-25M.} \cite{xie2024medtrinity} The dataset is a large-scale multimodal medical imaging dataset comprising approximately 25 million image--text triplets aggregated from over 30 public sources. It covers 10 imaging modalities and over 65 disease categories. Importantly, it provides annotations in an image-ROI-description format. We extract brain-related MRI samples, resulting in approximately 7 million brain MRI image-text pairs for pretraining.

\subsubsection{BRATS2023-GLI.} \cite{BRATS2023-GLI} The dataset comprises 1,251 multi-center, pre-operative glioma MRI studies from Sub-Saharan Africa, each with four modalities (T1n, T1c, T2w, T2f). We extract 2D slices centered on lesion masks along three axes from each 3D volume. Clinical reports are translated and structured into English using AutoRG Brain \cite{lei2024autorg} with GPT-4 prompts. Image-impression pairs are used for evaluation, with a 7:1:2 train/validation/test split.

\subsubsection{Metrics.}
For medical report generation, we adopt two widely used evaluation metrics: BLEU \cite{papineni2002bleu} and ROUGE-L \cite{lin2004rouge}. BLEU measures the n-gram precision of generated text against reference reports, while ROUGE-L evaluates the longest common subsequence between generated and reference texts, emphasizing recall.

\subsection{Implementation Details}
\subsubsection{Pretraining Stage.}
Following previous works \cite{khattak2024unimed,mu2022slip,zhang2023biomedclip}, we adopt ViT-B-16-quickgelu \cite{xu2023demystifying} as the visual encoder and BioMed-BERT \cite{chakraborty2020biomedbert} as the text encoder and momentum encoder, respectively. We train TextSLIP for 20 epochs on a single-node training setup with six A100 80GB GPUs. The learning rate is set to 4e-5 with a warm-up of 5k iterations, the effective batch size is configured as 1536, and the momentum coefficient is set to $\alpha=0.99$. The total training time for our model is approximately 80 hours.

\subsubsection{Fine-tuning Stage.} To evaluate downstream performance, we integrate our pretrained TextSLIP visual encoder into the R2Gen framework \cite{chen2020generatingr2gen}, replacing the original visual extractor. For comparison, we also incorporate visual encoders from two representative CLIP-style medical foundation models, UniMed-CLIP \cite{khattak2024unimed} and BioMed-CLIP \cite{zhang2023biomedclip}. All models undergo end-to-end fine-tuning on a single NVIDIA A100 80GB GPU for 30 epochs. We follow the default configuration specified in R2Gen for all other settings. During inference, embeddings are extracted from three anatomical axes, concatenated, and fed into the decoder module for report generation.

\subsection{Results}

In this section, we present experimental results organized into two parts. First, we compare TextSLIP against representative medical vision-language models. Second, we conduct ablation studies to analyze the impact of key components.

\subsubsection{Comparison against other language-aware visual encoders.} We first evaluated TextSLIP by replacing R2Gen's visual extractor with pretrained CLIP-style encoders (e.g., UniMed-CLIP \cite{khattak2024unimed}, BioMed-CLIP \cite{zhang2023biomedclip}). As shown in Table \ref{tab:Comparison}, TextSLIP achieves higher BLEU scores than these CLIP variants across MRI modalities T1n, T1c, T2w, and T2f. Compared with the strongest baseline in this comparison, UniMed-CLIP, TextSLIP obtains a gain of approximately 1--2\% in BL-1. Improvements are also observed on higher-order BLEU metrics, suggesting better generation of longer clinical descriptions under this evaluation setting.

\begin{table}[!b]
\caption{Comparison of different pretrained CLIP-style models. The best results are in \textbf{bold}.}
\label{tab:Comparison}
\centering
\begin{tabularx}{\linewidth}{p{0.10\linewidth}|p{0.18\linewidth}|*{5}{>{\centering\arraybackslash}X}}
\hline
Dataset & Model  & BL-1 & BL-2 & BL-3 & BL-4 & ROUGE-L\\
\hline
\multirow{3}{*}{T1n} 
    & TextSLIP & \textbf{0.6241} & \textbf{0.4827} & \textbf{0.3846} & \textbf{0.3068} & 0.5527 \\
    & UniMed-CLIP & 0.6052 & 0.4694 & 0.3762 & 0.2993 & \textbf{0.5543} \\
    & BioMed-CLIP & 0.6119 & 0.4755 & 0.3761 & 0.2971 & 0.5525 \\
\hline
\multirow{3}{*}{T1c} 
    & TextSLIP & \textbf{0.6332} & \textbf{0.4980} & \textbf{0.3942} & \textbf{0.3133} & \textbf{0.5668} \\
    & UniMed-CLIP& 0.6095 & 0.4778 & 0.3869 & 0.3027 & 0.5565 \\
    & BioMed-CLIP & 0.6087 & 0.4767 & 0.3799 & 0.2997 & 0.5576 \\
\hline
\multirow{3}{*}{T2w} 
    & TextSLIP & \textbf{0.6246} & \textbf{0.4814} & \textbf{0.3868} & 0.3103 & \textbf{0.5685} \\
    & UniMed-CLIP & 0.6139 & 0.4790 & 0.3863 & \textbf{0.3121} & 0.5620 \\
    & BioMed-CLIP & 0.6108 & 0.4788 & 0.3856 & 0.3086 & 0.5566 \\
\hline
\multirow{3}{*}{T2f} 
    & TextSLIP & \textbf{0.6372} & \textbf{0.5022} & \textbf{0.4125} & \textbf{0.3369} & \textbf{0.5810} \\
    & UniMed-CLIP & 0.6152 & 0.4826 & 0.3900 & 0.3124 & 0.5608 \\
    & BioMed-CLIP & 0.6239 & 0.4842 & 0.3857 & 0.3048 & 0.5456 \\
\hline
\end{tabularx}
\end{table}

\subsubsection{Ablation Study.} To isolate the impact of the pretraining objective, we conduct a controlled ablation study on the visual encoder. Specifically, we replace the R2Gen backbone with either a CLIP model pretrained on the same 7 million brain MRI pairs or the original unpretrained ResNet-101 trained from scratch following the standard R2Gen protocol. TextSLIP consistently surpasses both baselines across all modalities, with gains of approximately 3\% in BL-4 in several settings. These results suggest that text-side self-supervision combined with image-text contrastive learning can yield more discriminative visual representations for downstream medical report generation.

\begin{table}[!b]
\caption{\textbf{Ablation study on visual encoder initialization.} Comparison of TextSLIP, CLIP (pretrained on the same 7M brain MRI dataset), and the original unpretrained ResNet-101 backbone from R2Gen~\cite{chen2020generatingr2gen}. The best results are in \textbf{bold}.}
\label{tab:Ablation}
\centering
\begin{tabularx}{\linewidth}{p{0.10\linewidth}|p{0.18\linewidth}|*{5}{>{\centering\arraybackslash}X}}
\hline
Dataset & Model  & BL-1 & BL-2 & BL-3 & BL-4 & ROUGE-L \\
\hline
\multirow{3}{*}{T1n} 
    & TextSLIP & \textbf{0.6241} & \textbf{0.4827} & \textbf{0.3846} & \textbf{0.3068} & 0.5527 \\
    & CLIP & 0.6033 & 0.4659 & 0.3641 & 0.2845 & 0.5521 \\
    & ResNet101 & 0.5994 & 0.4660 & 0.3722 & 0.2969 & \textbf{0.5578} \\
\hline
\multirow{3}{*}{T1c} 
    & TextSLIP & \textbf{0.6332} & \textbf{0.4980} & \textbf{0.3942} & \textbf{0.3133} & \textbf{0.5668} \\
    & CLIP & 0.6095 & 0.4778 & 0.3869 & 0.3112 & 0.5596 \\
    & ResNet101 & 0.6049 & 0.4610 & 0.3631 & 0.2884 & 0.5411 \\
\hline
\multirow{3}{*}{T2w} 
    & TextSLIP & \textbf{0.6246} & \textbf{0.4814} & \textbf{0.3868} & \textbf{0.3103} & \textbf{0.5685} \\
    & CLIP & 0.5967 & 0.4616 & 0.3664 & 0.2927 & 0.5608 \\
    & ResNet101 & 0.5889 & 0.4584 & 0.3664 & 0.2910 & 0.5531 \\
\hline
\multirow{3}{*}{T2f} 
   & TextSLIP & \textbf{0.6372} & \textbf{0.5022} & \textbf{0.4125} & \textbf{0.3369} & \textbf{0.5810} \\
    & CLIP & 0.6030 & 0.4644 & 0.3741 & 0.2986 & 0.5579 \\
    & ResNet101 & 0.6106 & 0.4788 & 0.3885 & 0.3130 & 0.5583\\
\hline
\end{tabularx}
\end{table}

\section{Conclusion and Limitations}
In this work, we introduced TextSLIP, a text self-supervised pretraining framework for medical report generation. By integrating text contrastive learning with cross-modal image-text alignment, TextSLIP provides additional fine-grained semantic supervision to the visual encoder. Experiments on 7 million brain MRI image-text pairs show that TextSLIP improves over CLIP-style baselines under controlled training and evaluation settings. Ablation studies and qualitative analyses further suggest that structured textual embeddings can benefit downstream report generation. This study serves as an initial validation of TextSLIP: current experiments focus on brain MRI, evaluation primarily relies on lexical metrics such as BLEU and ROUGE-L, and the downstream reports are derived through a structured processing pipeline. Future work will extend TextSLIP to broader medical imaging domains, incorporate clinically grounded metrics and expert evaluation, and explore medical semantics-aware text augmentation that better preserves entities, negation, and uncertainty.

\bibliographystyle{splncs04}
\bibliography{ref}

@article{xie2024medtrinity,
  title={Medtrinity-25m: A large-scale multimodal dataset with multigranular annotations for medicine},
  author={Xie, Yunfei and Zhou, Ce and Gao, Lang and Wu, Juncheng and Li, Xianhang and Zhou, Hong-Yu and Liu, Sheng and Xing, Lei and Zou, James and Xie, Cihang and others},
  journal={arXiv preprint arXiv:2408.02900},
  year={2024}
}

@article{BRATS2023-GLI,
  title={The brain tumor segmentation (brats) challenge 2023: Glioma segmentation in sub-saharan africa patient population (brats-africa)},
  author={Adewole, Maruf and Rudie, Jeffrey D and Gbdamosi, Anu and Toyobo, Oluyemisi and Raymond, Confidence and Zhang, Dong and Omidiji, Olubukola and Akinola, Rachel and Suwaid, Mohammad Abba and Emegoakor, Adaobi and others},
  journal={ArXiv},
  pages={arXiv--2305},
  year={2023}
}

@article{lei2024autorg,
  title={Autorg-brain: Grounded report generation for brain mri},
  author={Lei, Jiayu and Zhang, Xiaoman and Wu, Chaoyi and Dai, Lisong and Zhang, Ya and Zhang, Yanyong and Wang, Yanfeng and Xie, Weidi and Li, Yuehua},
  journal={arXiv preprint arXiv:2407.16684},
  year={2024}
}

@inproceedings{mu2022slip,
  title={Slip: Self-supervision meets language-image pre-training},
  author={Mu, Norman and Kirillov, Alexander and Wagner, David and Xie, Saining},
  booktitle={European conference on computer vision},
  pages={529--544},
  year={2022},
  organization={Springer}
}

@inproceedings{radford2021learningclip,
  title={Learning transferable visual models from natural language supervision},
  author={Radford, Alec and Kim, Jong Wook and Hallacy, Chris and Ramesh, Aditya and Goh, Gabriel and Agarwal, Sandhini and Sastry, Girish and Askell, Amanda and Mishkin, Pamela and Clark, Jack and others},
  booktitle={International conference on machine learning},
  pages={8748--8763},
  year={2021},
  organization={PmLR}
}

@inproceedings{chen2020generatingr2gen,
  title={Generating radiology reports via memory-driven transformer},
  author={Chen, Zhihong and Song, Yan and Chang, Tsung-Hui and Wan, Xiang},
  booktitle={Proceedings of the 2020 conference on empirical methods in natural language processing (EMNLP)},
  pages={1439--1449},
  year={2020}
}

@inproceedings{chen2021crossr2genCMN,
  title={Cross-modal memory networks for radiology report generation},
  author={Chen, Zhihong and Shen, Yaling and Song, Yan and Wan, Xiang},
  booktitle={Proceedings of the 59th annual meeting of the association for computational linguistics and the 11th international joint conference on natural language processing (volume 1: long papers)},
  pages={5904--5914},
  year={2021}
}

@inproceedings{wang2023metransformer,
  title={Metransformer: Radiology report generation by transformer with multiple learnable expert tokens},
  author={Wang, Zhanyu and Liu, Lingqiao and Wang, Lei and Zhou, Luping},
  booktitle={Proceedings of the IEEE/CVF conference on computer vision and pattern recognition},
  pages={11558--11567},
  year={2023}
}

@inproceedings{li2023dynamic,
  title={Dynamic graph enhanced contrastive learning for chest x-ray report generation},
  author={Li, Mingjie and Lin, Bingqian and Chen, Zicong and Lin, Haokun and Liang, Xiaodan and Chang, Xiaojun},
  booktitle={Proceedings of the IEEE/CVF conference on computer vision and pattern recognition},
  pages={3334--3343},
  year={2023}
}

@inproceedings{wang2025cxpmrg,
  title={Cxpmrg-bench: Pre-training and benchmarking for x-ray medical report generation on chexpert plus dataset},
  author={Wang, Xiao and Wang, Fuling and Li, Yuehang and Ma, Qingchuan and Wang, Shiao and Jiang, Bo and Tang, Jin},
  booktitle={Proceedings of the computer vision and pattern recognition conference},
  pages={5123--5133},
  year={2025}
}

@inproceedings{yan2022clinical,
  title={Clinical-bert: Vision-language pre-training for radiograph diagnosis and reports generation},
  author={Yan, Bin and Pei, Mingtao},
  booktitle={Proceedings of the AAAI conference on artificial intelligence},
  volume={36},
  number={3},
  pages={2982--2990},
  year={2022}
}

@article{zhang2023biomedclip,
  title={Biomedclip: a multimodal biomedical foundation model pretrained from fifteen million scientific image-text pairs},
  author={Zhang, Sheng and Xu, Yanbo and Usuyama, Naoto and Xu, Hanwen and Bagga, Jaspreet and Tinn, Robert and Preston, Sam and Rao, Rajesh and Wei, Mu and Valluri, Naveen and others},
  journal={arXiv preprint arXiv:2303.00915},
  year={2023}
}

@article{khattak2024unimed,
  title={Unimed-clip: Towards a unified image-text pretraining paradigm for diverse medical imaging modalities},
  author={Khattak, Muhammad Uzair and Kunhimon, Shahina and Naseer, Muzammal and Khan, Salman and Khan, Fahad Shahbaz},
  journal={arXiv preprint arXiv:2412.10372},
  year={2024}
}

@inproceedings{papineni2002bleu,
  title={Bleu: a method for automatic evaluation of machine translation},
  author={Papineni, Kishore and Roukos, Salim and Ward, Todd and Zhu, Wei-Jing},
  booktitle={Proceedings of the 40th annual meeting of the Association for Computational Linguistics},
  pages={311--318},
  year={2002}
}

@inproceedings{lin2004rouge,
  title={Rouge: A package for automatic evaluation of summaries},
  author={Lin, Chin-Yew},
  booktitle={Text summarization branches out},
  pages={74--81},
  year={2004}
}

@inproceedings{chakraborty2020biomedbert,
  title={BioMedBERT: A pre-trained biomedical language model for QA and IR},
  author={Chakraborty, Souradip and Bisong, Ekaba and Bhatt, Shweta and Wagner, Thomas and Elliott, Riley and Mosconi, Francesco},
  booktitle={Proceedings of the 28th international conference on computational linguistics},
  pages={669--679},
  year={2020}
}

@article{xu2023demystifying,
  title={Demystifying clip data},
  author={Xu, Hu and Xie, Saining and Tan, Xiaoqing Ellen and Huang, Po-Yao and Howes, Russell and Sharma, Vasu and Li, Shang-Wen and Ghosh, Gargi and Zettlemoyer, Luke and Feichtenhofer, Christoph},
  journal={arXiv preprint arXiv:2309.16671},
  year={2023}
}

@inproceedings{wu2022esimcse,
  title={Esimcse: Enhanced sample building method for contrastive learning of unsupervised sentence embedding},
  author={Wu, Xing and Gao, Chaochen and Zang, Liangjun and Han, Jizhong and Wang, Zhongyuan and Hu, Songlin},
  booktitle={Proceedings of the 29th International Conference on Computational Linguistics},
  pages={3898--3907},
  year={2022}
}

@article{oord2018representation,
  title={Representation learning with contrastive predictive coding},
  author={Oord, Aaron van den and Li, Yazhe and Vinyals, Oriol},
  journal={arXiv preprint arXiv:1807.03748},
  year={2018}
}
\end{document}